\documentclass{article}
 
\usepackage[final]{neurips_2019-Sets}

\usepackage[utf8]{inputenc} 
\usepackage[T1]{fontenc} 
\usepackage{hyperref} 
\usepackage{url} 
\usepackage{booktabs} 
\usepackage{amsfonts} 
\usepackage{nicefrac} 
\usepackage{microtype} 

\usepackage{algorithm,algpseudocode}
\usepackage{amsmath}
\usepackage{amsfonts}
\usepackage{amssymb}

\usepackage{mathtools}

\DeclarePairedDelimiter\floor{\lfloor}{\rfloor}

\title{PairNets: Novel Fast Shallow Artificial Neural Networks on Partitioned Subspaces }

\author{Luna M. Zhang
}

\begin{document}

\maketitle

\begin{abstract}
Traditionally, an artificial neural network (ANN) is trained slowly by a gradient descent algorithm such as the backpropagation algorithm since a large number of hyperparameters of the ANN need to be fine-tuned with many training epochs. To highly speed up training, we created a novel shallow 4-layer ANN called ``Pairwise Neural Network" (``PairNet") with high-speed hyperparameter optimization. In addition, a value of each input is partitioned into multiple intervals, and then 
an $n$-dimensional space is partitioned into $M$ $n$-dimensional subspaces. $M$ local PairNets are built in $M$ partitioned local $n$-dimensional subspaces. 
A local PairNet is trained very quickly with only one epoch since its hyperparameters are directly optimized one-time via simply solving a system of linear equations by using the multivariate least squares fitting method. Simulation results for three regression problems indicated that the PairNet achieved much higher speeds and lower average testing mean squared errors (MSEs) for the three cases, and lower average training MSEs for two cases than the traditional ANNs. A significant future work is to develop better and faster optimization algorithms based on intelligent methods and parallel computing methods to optimize both partitioned subspaces and hyperparameters to build the fast and effective PairNets for applications in big data mining and real-time machine learning.
\end{abstract}

\section{Introduction}

Traditionally, an artificial neural network (ANN) is trained very slowly by a gradient descent algorithm such as the backpropagation algorithm [1-3] since a large number of hyperparameters of the ANN need to be fine-tuned with a larger number of training epochs. In particular, a deep neural network [4-9], such as a convolutional neural network (CNN), typically takes a long time to be trained well. Other intelligent training algorithms use various advanced optimization methods such as genetic algorithms [10-17], particle swarm optimization methods [18], and annealing algorithms [19] to try to find optimal hyperparameters of an ANN. However, these commonly used training algorithms take very long training time. An important research goal is to develop a new ANN with high computation speed and high performance, such as low validation errors, for various machine learning applications especially involving big data mining and real-time computation.

Neural network structure optimization algorithms also take a lot of time to try to find optimal or near-optimal numbers of different layers and numbers of neurons on different layers for big data mining problems. Especially, deep neural networks need much longer time. Thus, it is useful to develop fast shallow neural networks with relatively small numbers of neurons on different layers. We created a novel shallow 4-layer ANN with high-speed hyperparameter optimization. Training data are partitioned into local $n$-dimensional subspaces. Local shallow 4-layer ANNs are trained by using the partitioned data sets in the local $n$-dimensional subspaces. This divide-and-conquer approach can optimize the local ANNs with simpler nonlinear functions more easily using smaller data sets in the local subspaces. Based on positive preliminary simulation results, we will continue to develop more advanced optimization algorithms to optimize both partitioned subspaces and hyperparameters to build fast and effective ANNs.

\section{Pairwise Neural Network (PairNet)}

We propose a novel shallow ANN called “Pairwise Neural Network” (PairNet) that consists of only four layers of neurons to map $n$ inputs on the first layer to one output on the fourth layer. 

{\em Layer 1}: Layer 1 has $n$ neuron pairs to map $n$ inputs to $2n$ outputs. Each pair has two neurons where one neuron has an increasing activation function $g_{i}(x_{i})\in[0,1]$ that generates a positive normalized value, and the other neuron has a decreasing activation function $(1-g_{i}(x_{i}))$ that generates a negative normalized value for $i=1,2,...,n$. 

{\em Layer 2}: Layer 2 consists of $2^n$ neurons, where each neuron has an activation function to map $n$ inputs to an output as a complementary decision fusion. Each of the $n$ inputs is an output of one of the two neurons of each neuron pair on Layer 1. Let $g_{i}$ denote $g_{i}(x_{i})$, and $\bar{g}_{i}$ denote $(1-g_{i}(x_{i}))$ for $i=1,2,...,n$. The activation functions of neurons on Layer 2 are given as $w_{1}
=\alpha_{1}g_1+\alpha_{2}g_2+...+\alpha_{n-1}g_{n-1}+\alpha_{n}g_n$, ..., $w_{2^{n}-1}=\alpha_{1}\bar{g}_1+\alpha_{2}\bar{g}_2+...+\alpha_{n-1}\bar{g}_{n-1}+\alpha_{n}g_n$, $w_{2^{n}}
=\alpha_{1}\bar{g}_1+\alpha_{2}\bar{g}_2+...+\alpha_{n-1}\bar{g}_{n-1}+\alpha_{n}\bar{g}_n$, where $\alpha_{i}$ are hyperparameters to be optimized for $0\leq \alpha_{i}\leq 1$, $i=1,2,...,n$, and $\sum_{i=1}^{n}\alpha_{i}=1$. $\sum_{k=1}^{2^{n}}w_k=2^{n-1}\sum_{i=1}^{n}\alpha_{i}(g_i+\bar{g}_i)
=2^{n-1}\sum_{i=1}^{n}\alpha_{i}=2^{n-1}$. For a special case, the weights ($\alpha_{i}$) are equal, so $\alpha_{i}=\frac{1}{n}$ for $i=1,2,...,n$.

{\em Layer 3}: Layer 3 also consists of $2^n$ neurons but transforms the outputs of the second layer to $2^n$ individual output decisions. $w_k=1+\frac{y_{k}^{1}-c_{k}}{\eta_{k}}$ for $ (c_{k}-\eta_{k})
\leq y_{k}\leq c_{k}$, $w_k=1- \frac{y_{k}^{2}-c_{k}}{\delta_{k}}$ for 
$ c_{k}
\leq y_{k}\leq (c_{k}+\delta_{k})$, where $k=1,2,...,2^{n}$. $\bar{y}_k$ (activation functions) are defined as $\bar{y}_k = \frac{y_{k}^{1} + y_{k}^{2}}{2} = c_{k} + \frac{(1 - w_k)\gamma_k}{2}$, where $\gamma_k = \delta_k - \eta_k$. 

{\em Layer 4}: Layer 4 generates a final nonlinear output $y=f(\bar{y}_1,\bar{y}_2,...,\bar{y}_{2^{n}})$.

\section{Fast Training Algorithm with Hyperparameter Optimization on Partitioned Subspaces}

A data set has $N$ data, where each data consists of $n$ inputs $x_i$ for $i=1,2,...,n$, and one output $y$. An input $x_i$ has $m_i$ intervals in $[a_i, b_i]$ such that 
$[a_i, a_{i1}]$, $[a_{i1}, a_{i2}]$, ..., $[a_{im_{i}-2}, a_{im_{i}-1}]$, and $[a_{im_{i}}, b_i]$ for $m_i\geq1$, and $i=1,2,...,n$. Then there are $M$
($M=\prod_{i=1}^{n}m_i$) $n$-dimensional subspaces $S_{j}$ for $j=1,2,...,M$. $N$ data are distributed in the $M$ $n$-dimensional subspaces. A $n$-dimensional subspace $S_{j}$ has $N_{j}$ data with $N_{j}$ outputs $Y^j_p$ for $j=1,2,...,M$, $p=1,2,...,N_{j}$, and $N=\sum_{j=1}^{M}N_{j}$. 
For each $n$-dimensional subspace such as ($[a_{11}, a_{12}]$, $[a_{21}, a_{22}]$, ..., $[a_{n-11}, a_{n-12}]$, and $[a_{n1}, a_{n2}]$), 
a PairNet can map $n$ inputs $x_i$ for $i=1,2,...,n$ to one output $f_{j}(x_{1},...,x_{n})$ for $j=1,2,...,M$. Thus, a local PairNet is built using all of the data points in a local $n$-dimensional subspace. This divide-and-conquer approach can train the local PairNet using specific local data features to improve model performance. 

For a regression problem, Layer 4 of a PairNet
calculates a final output decision by computing a weighted average of the $2^n$ individual output decisions of Layer 3. 
The final ouptut is generated by a nonlinear function $f(x_1,x_2,...,x_n)=\sum_{k=1}^{2^{n}}
\beta_k \bar{y}_k$, , where $\beta_k = \frac{w_k}{\sum_{j=1}^{2^{n}}w_j}=\frac{w_k}{2^{n-1}}$. Finally, $f(x_1,x_2,...,x_n)= \sum_{k=1}^{2^{n}}(\beta_{k}c_k + \beta_{k}\theta_{k}\gamma_k)$, where $\theta_{k}=\frac{1 - w_{k}}{2}$ for $k=1,2,...,2^{n}$. 

The objective optimization function for a PairNet $f_{j}(x_{1},...,x_{n})$ for $j=1,2,...,M$ is given below,
\begin{equation}
Q=\frac{1}{2}\sum_{p=1}^{N_{j}}[Y^j_p-f_{j}(x_{1_p},x_{2_p},...,x_{n_p})]^2.\
\end{equation}

After setting $\frac{\partial Q}{\partial c^j_k}=0$ and $\frac{\partial Q}{\partial \gamma^j_k}=0$, we have $2^{n+1}$ linear equations with $2^{n+1}$ hyperparameters ($c_k$ and $\gamma_k$) for $k=1,2,...,2^{n}$ as follows:
{
\begin{eqnarray}
\left\{\begin{array}{lll}
\sum_{p=1}^{N}\beta^j_{k_p}(Y^j_p - \sum_{q=1}^{2^{n}}(\beta^j_{q_p}c_q + \beta^j_{q_p}\theta^j_{q_p}\gamma^j_q))
=0\\
\sum_{p=1}^{N}\beta^j_{k_p}\theta^j_{k_p}(Y^j_p - \sum_{q=1}^{2^{n}}(\beta^j_{q_p}c_q + \beta^j_{q_p}\theta^j_{q_p}\gamma^j_q))
=0
.\\
\end{array}
\right.
\end{eqnarray}
}

The above system of linear equations (2) can  be  quickly solved to find optimal hyperparameters ($c_k$ and $\gamma_k$) for $k=1, 2,..., 2^{n}$. Each subspace must have at least $2^{n+1}$ data points.
A new PairNet model selection algorithm is given in Algorithm 1.\\

\begin{algorithm}[H] 
\caption{ PairNet Model Selection Algorithm with Fast Hyperparameter Optimization}
\label{alg:loop}
\begin{algorithmic}[1]
\Require{$K$: the number of candidate PairNet models} 
\Ensure{the best PairNet model} 

\State {Randomly generate $M$ subspaces $S_j$ 
for $j=1,2,...,M$. }
\State {Calculate hyperparameters using equation (2) for $M$ subspaces to generate $M$ local PairNets.}
\State {Evaluate the performance of the $M$ local PairNets.}
\State {Set the best model to be this PairNet, which has $M$ local PairNets.}

\For{$k = 1$ to $K$} 
\State {Randomly generate $M_k$ subspaces $S^k_j$ for $j=1,2,...,M_k $. }
\State {Calculate hyperparameters using equation (2) for $M$ subspaces to generate $M_k$ local PairNets.}
\State {Evaluate the performance of the $M_k$ local PairNets.}
\State {If this newly generated PairNet (which has $M$ local PairNets) has better performance than the best PairNet, then the best model is the newly generated PairNet.}
\EndFor
\State \Return {the best PairNet model.}

\end{algorithmic}
\end{algorithm}

\section{Simulation Results}

To compare an ANN and the PairNet, three different simulations using three different functions are 
done. The first 3-input-1-output benchmark function [20-23] is given below:
{
\begin{eqnarray}
f^1_k=(1+x_{k}^{0.5}+y_{k}^{-1}+z_{k}^{-1.5})^2.\
\end{eqnarray} 
}
The second 3-input-1-output function is given below:
{
\begin{eqnarray}
f^2_k=x_{k}^{1.25}sin(x_{k}^{0.15}-z_{k}^{0.05})+y_{k}^{1.25}+z_{k}^{0.15}.\
\end{eqnarray} 
}
The third 3-input-1-output function is given below:
{
\begin{eqnarray}
f^3_k=(1+x_{k}^{0.25}z_{k}+y_{k}^{0.5}z_{k}^{-1}+z_{k}^{-0.05})^2.\
\end{eqnarray} 
}Three training data sets (each with $8000$ training data) are generated by the three functions 
shown in equations (3), (4), and (5) such that 
$x^{tr}_{k}=1.0+\floor{\frac{k}{400}}$,
$y^{tr}_{k}=1.0+\floor{\frac{k}{20}}$, 
$z^{tr}_{k}=1.0+k $ $mod20$, 
where the operator $mod$ is used, $f^1_k \in [4.248, 55.833]$, $f^2_k \in [2.0, 66.023]$, $f^3_k\in [16.0, 1969.527]$, and $k=0,1,...,7999$. Three testing data sets (each with $6859$ testing data) are generated 
by the three functions 
shown in equations (3), (4), and (5) such that 
$x^{te}_{j}=1.5+\floor{\frac{j}{361}}$,
$y^{te}_{j}=1.5+\floor{\frac{j}{19}}$, 
$z^{te}_{j}=1.5+j$ $mod19$, 
where the operator $mod$ is used, $j=0,1,...,6858$. 

For simulations, the best 20-layer ANN was selected from five random 20-layer ANNs using ReLU (500 epochs), and the best PairNet was selected from five random PairNets with random $\alpha_{i}$ for $i=1,2,3$ and random 3-dimensional subspaces. Results shown in Table 1 indicated that the PairNets outperformed traditional ANNs in terms of speed and testing mean squared errors (MSEs).

\begin{table}[h!]
\caption{Performance comparison between the best ANNs and the best PairNets}
\label{sample-table}
\centering
\begin{tabular}{lllllllll}
\toprule
\bf{Method} & \bf{Function} & \bf{$\bf{T_{train}}$ (sec)} & $\bf{MSE_{train}}$ & $\bf{MSE_{test}}$ \\
\midrule
PairNet& $f^1$ & \textbf{3.06} & 0.191 & \textbf{0.225} \\
ANN& $f^1$ & 199.3 & \textbf{0.022} & 0.249 \\
\hline
PairNet& $f^2$ & \textbf{3.01} & \textbf{0.00075} & \textbf{0.00227} \\
ANN& $f^2$ & 192.6 & 0.04214 & 0.02510 \\
\hline
PairNet& $f^3$ & \textbf{2.84} & \textbf{10.930} & \textbf{7.7513} \\
ANN& $f^3$ & 277.6 & 86.861 & 66.798 \\
\bottomrule
\end{tabular}
\end{table}

Simulation results for $\alpha_{1}=0.1$, $\alpha_{2}=0.1$, and $\alpha_{3}=0.8$ shown in Table 2 indicated that the more number of different partitioned subspaces,
the better a PairNet tended to perform in terms of training MSE ($MSE_{tr}$) and testing MSE ($MSE_{te}$) in most cases. In addition, a PairNet with more subspaces is not always better than that with fewer  subspaces. An important future work is to develop a new high-speed optimization algorithm to find both best partitioned subspaces and optimal hyperparameters for building the best PairNet.

\begin{table}[h!]
\caption{Performance analysis for the PairNets on different partitioned subspaces for $f^1$, $f^2$, and $f^3$}
\label{sample-table}
\centering
\begin{tabular}{lllllllll}
\toprule
\small {\bf{Partitions ($x$-$y$-$z$)}} & \small {\bf{Subspaces}} & \small {$\bf{{MSE}^{f^1}_{tr}}$} & \small {$\bf{{MSE}^{f^1}_{te}}$} & \small {$\bf{{MSE}^{f^2}_{tr}}$} & \small {$\bf{{MSE}^{f^2}_{te}}$} & \small {$\bf{{MSE}^{f^3}_{tr}}$} & \small {$\bf{{MSE}^{f^3}_{te}}$}\\
\midrule
2-2-2& 8 & 1.926 & 0.940 & 0.1713 & 0.1325 & 258.0 & 148.4 \\
2-3-4& 24 & 0.857 & 0.673 & 0.1091 & 0.0903& 224.3 & 132.4 \\
3-3-3& 27 & 0.939 & 0.606 & 0.0348 & 0.0302 & 78.30 & 46.39 \\
3-4-5& 60 & 0.444 & 0.624 & 0.0253 & 0.0242& 82.60 & 44.47 \\
4-4-4& 64 & 0.534 & 0.702 & 0.0111 & 0.0160& 37.60 & 37.60 \\
4-5-6& 120 & 0.245 & 0.426 & 0.0065 & 0.0122& 25.94 & 35.82 \\
5-5-5& 125 & 0.291 & 0.563 & 0.0041 & 0.0085 & 14.39 & 23.27 \\
6-6-6& 216 & 0.168 & 0.245 & 0.0018 & 0.0030& 7.966 & 7.300\\
\bottomrule
\end{tabular}
\end{table}

\section{Conclusions}

The new shallow 4-layer PairNet can be trained very quickly with only one epoch since its hyperparameters are directly optimized one-time via simply solving a system of linear equations by using the multivariate least squares fitting method. Different from gradient descent training algorithms and other training algorithms such as genetic algorithms, the new training algorithm with direct hyperparameter computation can quickly train the PairNet because it does not need slow training with a large number of epochs. Initial simulation results show that the shallow PairNet is much faster than traditional ANNs. For accuracy, the PairNet may not always achieve the lowest training MSE but can achieve lower testing MSEs than traditional ANNs. 

In addition, the divide-and-conquer approach used by Algorithm 1 is effective and efficient to build local PairNet models on local $n$-dimensional subspaces. For big data mining applications, partitioning a big data space into many small data subspaces is useful since each local PairNet covering a small data subspace is built more quickly using fewer data points than a global PairNet covering the whole big data space. 

\section{Future Works}

More robust simulations with much more complex data sets with more inputs will be done to further evaluate the PairNet and to further compare the PairNet and traditional ANNs. Additionally, a new PairNet with a new activation function of the neuron on Layer 4 will be created for classification applications. The new PairNet will be further evaluated by commonly used benchmark classification problems. The PairNet can be optimized to reduce the training MSE and testing MSE by model selection via optimizing partitioned local $n$-dimensional subspaces. 

Although the PairNet is a shallow neural network, it is actually a wide neural network if $n$ is large because both the second layer and the third layer have $2^n$ neurons with the first layer having $n$ neurons. Thus, the PairNet has the curse of dimensionality. However, we will develop advanced divide-and-conquer methods to solve it. The preliminary simulations applied a random data partitioning method to divide a whole $n$-dimensional space into many $n$-dimensional subspaces. More intelligent data partitioning methods will be created to build more effective local PairNets on optimized $n$-dimensional subspaces. 

A significant future work is to develop more effective and faster hyperparameter optimization algorithms using parallel computing methods to find the best high-speed PairNet model with ideal activation functions on optimized $n$-dimensional subspaces for various applications in real-time machine learning and big data mining.

\section*{References}

\medskip

\small

[1] Werbos, P. (1974) Beyond Regression: New Tools for Prediction and Analysis in the Behavioral Sciences. PhD thesis, Harvard University. 

[2] Werbos, P. (1990) Backpropagation through time: what it does and how to do it. Proceedings of the IEEE 78(10): 1550--1160.

[3] Rumelhart, D. E.; Hinton, G. E.; \ \& Williams, R. J. (1986) Learning representations by back-propagating errors. Nature 323: 533--536.

[4] LeCun, Y., Bengio, Y.\ \& Hinton, G.E. (2015) Deep learning. Nature 521, pp.\ 436--444.

[5] Krizhevsky, A., Sutskever, I.\ \& Hinton, G.E.\ (2012) Imagenet classification with deep convolutional neural networks. In Advances in Neural Information Processing Systems 25, pp.\ 1097--1105. Cambridge, MA: MIT Press.

[6] He, K., Zhang, X., Ren, S.\ \& Sun, J. \ (2016) Deep Residual Learning for Image Recognition. In Proceedings of the 2016 IEEE Conference on Computer Vision and Pattern Recognition (CVPR), pp.\ 770--778.

[7] Esteva, A., Kuprel, B., Novoa, R.A., Ko, J., Swetter, S.M., Blau, H.M.\ \& Thrun, S.\ (2017) Dermatologist-level classification of skin cancer with deep neural networks. Nature 542(7639):115--118.

[8] Szegedy, C., Liu, W., Jia, Y., Sermanet, P., Reed S., Anguelov D., Erhan, D., Vanhoucke, V.\ \& Rabinovich, A.\ (2015) Going Deeper with Convolutions. 2015 IEEE Conference on Computer Vision and Pattern Recognition (CVPR), pp.\ 1--9. 

[9] Szegedy, C., Ioffe, S., Vanhoucke, V.\ \& Alemi, A.\ (2017) Inception-v4, Inception-ResNet and the Impact of Residual Connections on Learning. In Proceedings of the Thirty-First AAAI Conference on Artificial Intelligence (AAAI-17), pp.\ 4278--4284.

[10] Sun, Y., Xue, B., Zhang, M. \ \& Yen, G.\ (2018) Automatically Designing CNN Architectures Using Genetic Algorithm for Image Classification. [Online.] Available: $https://arxiv.org/pdf/1808.03818.pdf$.

[11] You, Z. \ \& Pu, Y. \ (2015) The Genetic Convolutional Neural Network Model Based on Random Sample. International Journal of u- and e- Service, Science and Technology, pp.\ 317--326.

[12] Ijjina, E. P. \ \& Mohan, C. K. (2016) Human action recognition using genetic algorithms and convolutional neural networks. Pattern Recognition, vol. 59, pp.\ 199--212.

[13] Fujino, S., Hatanaka, T., Mori, N. \ \& Matsumoto, K. \ (2017) The evolutionary deep learning basedon deep convolutional neural network for the anime storyboard recognition. International Symposium on Distributed Computing and Artificial Intelligence (DCAI 2017), pp.\ 278--285.

[14] Bochinski, E., Senst, T. \ \& Sikora, T. \ (2017) Hyper-parameter optimization for convolutional neural network committees based on evolutionary algorithms. 2017 IEEE International Conference on Image Processing (ICIP 2017), pp.\ 3924—3928. 

[15] Tian, H., Pouyanfar, S., Chen, J., Chen, S.-C. \ \& Iyengar, S. S. \ (2018) Automatic Convolutional Neural Network Selection for Image Classification Using Genetic Algorithms. 2018 IEEE International Conference on Information Reuse and Integration (IRI 2018), pp.\ 444-- 451. 

[16] Loussaief, S. \ \& Abdelkrim, A. \ (2018) Convolutional Neural Network Hyper-Parameters Optimization based on Genetic Algorithms. International Journal of Advanced Computer Science and Applications, vol. 9, no. 10, pp.\ 252—266.

[17] Baldominos, A., Saez, Y. \ \& Isasi, P. \ (2018) Model Selection in Committees of Evolved Convolutional Neural Networks using Genetic Algorithms. Intelligent Data Engineering and Automated Learning – IDEAL 2018, pp.\ 364--373. 

[18] Sinha, T., Haidar, A. \ \& Verma, B. \ (2018) 
Particle Swarm Optimization Based Approach for Finding Optimal Values of Convolutional Neural Network Parameters. 2018 IEEE Congress on Evolutionary Computation (CEC), pp.\ 1--6.

[19] Ayumi, V., Rasdi Rere, L. M., Fanany, M. I. \ \& Arymurthy, A. M. 
\ (2016) Optimization of convolutional neural network using microcanonical annealing algorithm. 2016 International Conference on Advanced Computer Science and Information Systems (ICACSIS), pp.\ 506--3511.

[20] Kondo T.\ (1986) Revised GMDH algorithm estimating degree of the complete polynomial. Trans. SOC. Instrument and Contr. Engineers, vol. 22, no. 9, pp. 928--934.

[21] Sugeno M. \& Kang G. T.\ (1988) Structure identification of fuzzy model, Fuzzy Sets Syst., vol. 28, pp. 15--33.

[22] Takagi H.\ \& Hayashi I.\ (1991) NN-driven fuzzy reasoning. Int. J. Approxi- mate Reasoning, vol. 5, no. 3, pp. 191--212.

[23] Jang, J.S.R. (1993) ANFIS: adaptive-network-based fuzzy inference system. \textit{IEEE Transactions on Systems, Man, and Cybernetics} 23(3): 665--685.

\end{document}